\documentclass[cameraready]{Interspeech}

\title{TASTE-\textit{Streaming}: Towards Streamable Text-Aligned Speech Tokenization and Embedding for Spoken Language Modeling}

\author[affiliation={1}, orcid=0009-0003-4084-2827]{Liang-Hsuan}{Tseng}
\author[affiliation={1}, orcid=0000-0002-9654-5747, correspondingauthor]{Hung-yi}{Lee}

\address{
    $^1$ Graduate Institute of Communication Engineering, National Taiwan University
}

\email{andybi7676@gmail.com, hungyilee@ntu.edu.tw}


\usepackage{comment}

\usepackage{adjustbox}
\usepackage{multicol}
\usepackage{multirow}
\usepackage{booktabs}
\usepackage{xcolor}
\usepackage{graphicx}
\usepackage[table]{xcolor}

\newcommand{\vct}[1]{\bm{#1}}   

\begin{document}

\maketitle

\begin{abstract}
Text-speech joint spoken language modeling (SLM) aims at natural and intelligent speech-based interactions, but developing such a system may suffer from modality mismatch: speech unit sequences are much longer than text tokens. 
Prior work reduces this gap with text-aligned tokenization and embedding (TASTE), producing speech tokens that align in lengths with their textual counterparts. 
However, the dependence on an external ASR system and the use of a non-causal decoder limits streaming use. 
To address this limitation, we propose TASTE-\textit{S}, a streamable extension of TASTE suitable for real-time usage. 
TASTE-\textit{S} integrates a CTC-based ASR module into the encoder for instant dual-modality encoding. 
We also redesign the unit decoder to enable on-the-fly decoding. 
With joint training, we show that TASTE-\textit{S} matches TASTE's performance while significantly reducing latency. 
Further investigations reveal that TASTE-\textit{S} remains robust to transcriptions and enables long-form encoding and decoding.
\end{abstract}

\section{Introduction}

Spoken Language Modeling (SLM) has been a promising direction of creating intelligent and natural human-machine interfaces. 
Conventional methods may develop such system by modeling speech tokens derived from self-supervised models~\cite{hubert, wav2vec2, gslm} or audio codec~\cite{encodec, soundstream}. 
However, modeling only speech tokens suffers from poor semantic grounding~\cite{dgslm, twist, spiritlm}.
To address this, recent research has pivoted toward \emph{joint text-speech modeling}, taking advantages of the understanding and reasoning capabilities from text Large Language Models (LLMs)~\cite{moshi, mini-omni, glm4-voice}.

\begin{figure}[t]
    \centering
    \includegraphics[width=\linewidth]{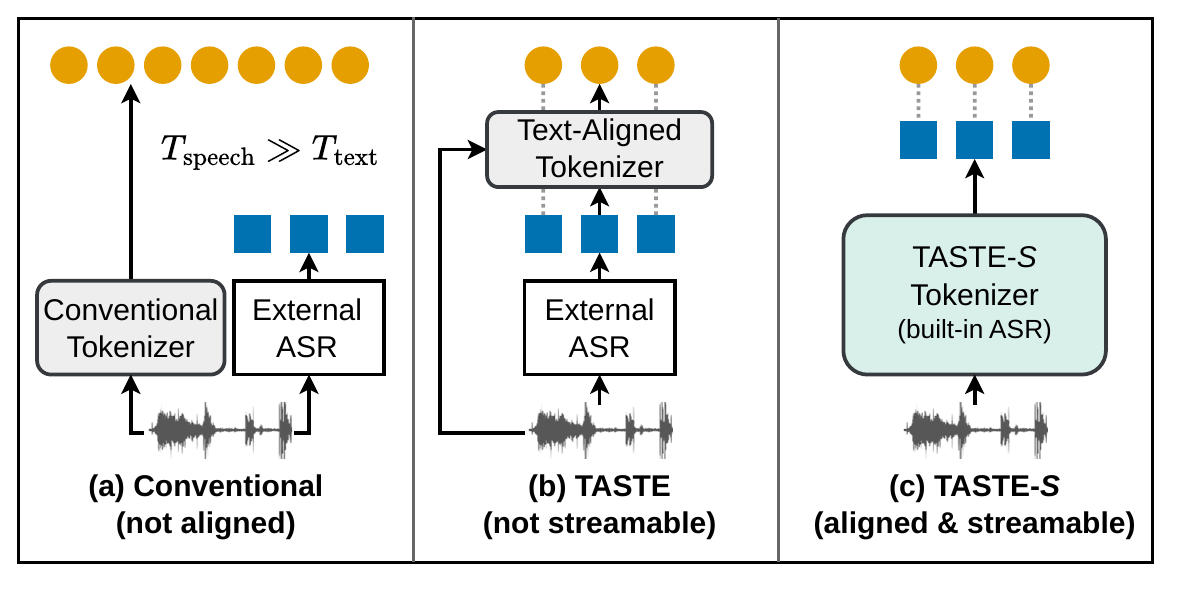}
    \caption{
    Comparison between conventional, TASTE and TASTE-\textit{S} speech tokenizer. 
    (a) conventional pipelines suffer from modality mismatch. 
    (b) TASTE aligns speech tokens to text tokens but is non-streamable since an external offline ASR is required. 
    (c) Our TASTE-\textit{S} achieves aligned \emph{and} streamable tokenization via a built-in ASR and other designs for streaming.
    }
    \label{fig:taste_s_overview}
\end{figure}

In joint text-speech modeling, a primary obstacle is the mismatch between speech and text. 
Conventional speech units are typically much finer-grained and longer than text tokens, leading to a significant length discrepancy (Figure~\ref{fig:taste_s_overview}-(a)). While prior work has attempted to resolve this through padding~\cite{llama-omni}, interleaving~\cite{spiritlm}, or multi-channel modeling~\cite{moshi}, these solutions often introduce additional complexity and efforts during joint modeling. 
The TASTE (Text-Aligned Tokenization and Embedding;~\cite{taste}) framework addressed this by mitigating the length mismatch issue at the tokenization stage (Figure~\ref{fig:taste_s_overview}-(b)), enabling simple and effective joint modeling~\cite{tasla, ddtaste}. 

Despite the success of text-aligned representations in boosting semantic performance, the original TASTE framework suffers from a critical drawback: latency. Because the alignment process was primarily designed for offline processing, it remains unsuitable for the "always-on," instant-response requirements of real-world conversational AI.

In this work, we present TASTE-\textit{S}, a streamable extension of the TASTE framework that provides text-aligned speech tokenization and embedding with low latency. 
It retains the semantic benefits of aligned tokenization while making on-the-fly encoding and reconstruction practical (Figure~\ref{fig:taste_s_overview}-(c)). 
Our modifications are three-fold:
\begin{itemize}
\item Integrated ASR Module: We seamlessly integrate an Automatic Speech Recognition (ASR) module into the tokenization system by attaching a lightweight Connectionist Temporal Classification (CTC) decoder to the speech encoder, facilitating immediate text token extraction.

\item Causal and Streamable Architecture: We redesign the unit decoder to be fully causal, allowing for low-latency, chunk-by-chunk processing without sacrificing the quality of the learned representations.

\item Bi-Stage and Joint Training: We show that with simple bi-stage training and the joint training between the Encoder and the Decoder, TASTE-\textit{S} Tokenizer can achieve higher and more stable performance across different textual transcripts.
\end{itemize}

Our experiments demonstrate that TASTE-\textit{S} achieves performance parity with the original framework while being significantly more efficient. 
Furthermore, we show that our tokenization is robust to varied textual transcriptions and capable of accurate long-form encoding and decoding. 
By visualizing the cross-attention maps, we confirm that the desired text-speech alignment is maintained. 
Ultimately, TASTE-\textit{S} provides a practical path toward responsive, semantically-aware Spoken Language Models. 
The demo is available at 
\url{https://andybi7676.github.io/taste_s_demo}

\section{Method}
TASTE-\textit{S} Tokenizer is composed of the two major components: The TASTE-\textit{S} Encoder that turns speech signal into text and text-aligned speech tokens and the TASTE-\textit{S} Decoder which takes them as the condition for speech reconstruction.\footnote{For clarity, Encoder/Decoder (capitalized) refer exclusively to the TASTE-\textit{S} Encoder/Decoder; whereas the lowercase suffixes (\(\mathrm{\_encoder}\), \(\mathrm{\_decoder}\)) denote the submodules, identifying their functionalities.}
We first introduce the TASTE-\textit{S} Encoder and Decoder in Section~\ref{subsec:taste_s_encoder} and Section~\ref{subsec:taste_s_decoder}, focusing on how to make the modules streamable. 
Then, we elaborate the objectives and our training procedure of the whole framework in Section~\ref{subsec:obj_and_training_proc}. 

\begin{figure}[t]
    \centering
    \includegraphics[width=1\linewidth]{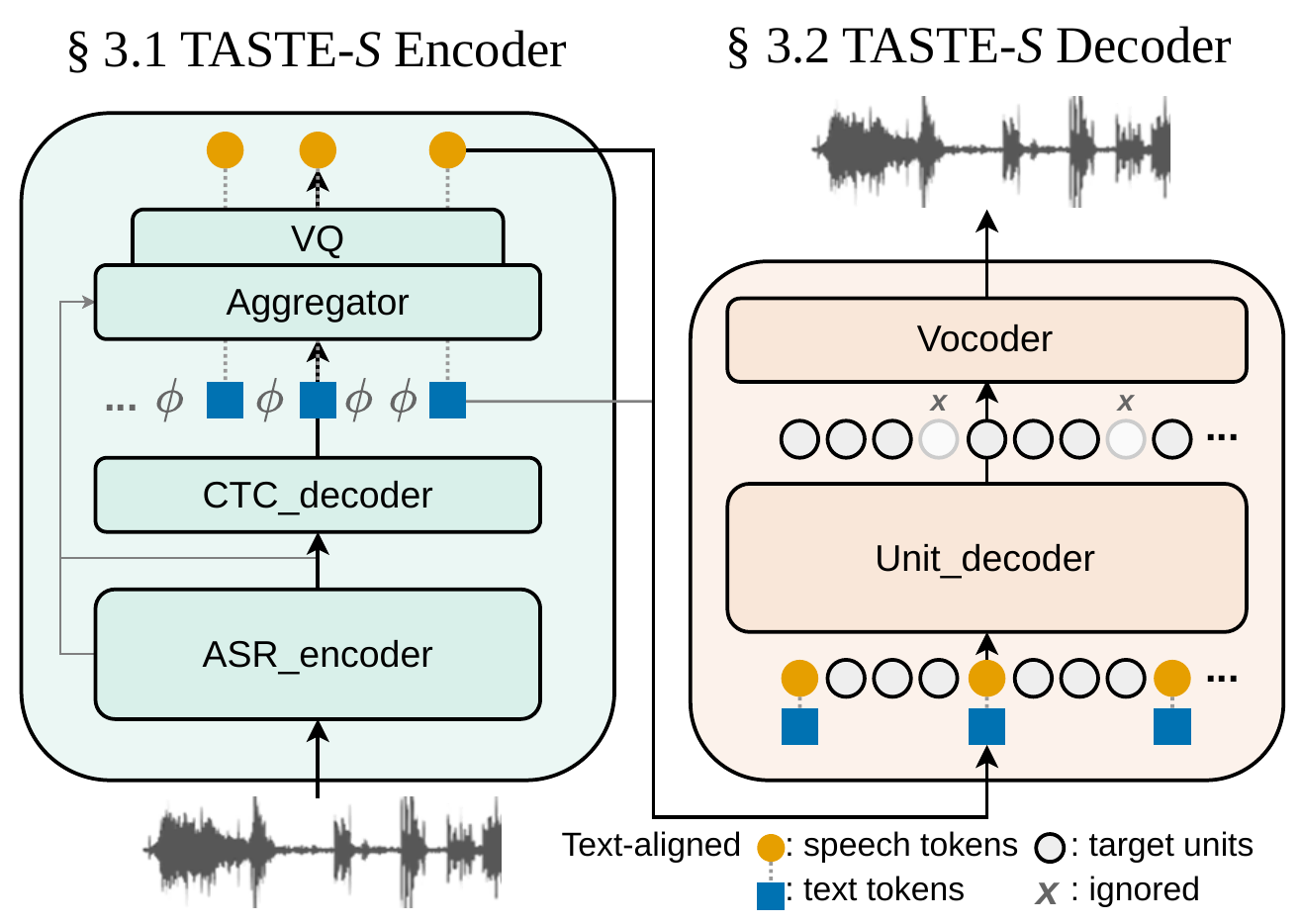}
    \caption{
    The framework overivew of our TASTE-\textit{S} Tokenizer. 
    On the left shows the Encoder with CTC integrated; on the right illustrates the streaming pattern for a streamable Decoder. 
    }
    \label{fig:taste_s_method}
\end{figure}

\subsection{TASTE-\textit{S} Encoder}
\label{subsec:taste_s_encoder}
The goal of our TASTE-\textit{S} Encoder is to encode raw speech into text-aligned speech tokens without introducing additional latency. 
Unlike TASTE~\cite{taste} that relies on an external offline ASR system, we internalize an ASR system into TASTE-\textit{S} Encoder. 
As depicted in Figure~\ref{fig:taste_s_method}, we adopt Connectionist Temporal Classification (CTC;~\cite{ctc}) to train our internalized ASR. 
This allows us to extract textual transcriptions as text tokens in an efficient and streamable manner. 
The TASTE-\textit{S} Encoding procedure is as follows:
Given a speech signal \(X\), we extract the hidden representations from the \(\mathrm{ASR\_encoder}\) with \(L\) layers, where
\begin{align}
    [\mat{H}^{(1)}, \mat{H}^{(2)}, \ldots, \mat{H}^{(L)}] = \mathrm{ASR\_encoder}(X).
\end{align}
Then, the \(\mathrm{CTC\_decoder}\) takes the last hidden representation \(\mat{H}^{(L)}\) as input and generates the ASR prediction logits \(\mat{O}\), which are then decoded into \(\vct{\hat{s}}\), denoted as:
\begin{align}
    \mat{O} &= \mathrm{CTC\_decoder}(\mat{H}^{(L)}), 
    \quad \vct{\hat{s}} = \mathrm{Decode_{CTC}}(\mat{O})
\end{align}
Next, following the original TASTE, we extract the text-aligned speech tokens through an \(\mathrm{Aggregator}\) followed by a vector quantizer \(\mathrm{VQ}\), with the predicted text tokens \(\vct{\hat{s}}\), shallow and last hiddens \(\mat{H}^{(l)}, \mat{H}^{(L)}\) from the encoder taken as input:
\begin{align}
     \mat{Z} = \mathrm{Aggregator}(\vct{\hat{s}}, \mat{H}^{(L)}, \mat{H}^{(l)}), \hspace{0.5em} \mat{\hat{Z}} = \mathrm{VQ}(\mat{Z}). 
\end{align}
Note that \(\mat{\hat{Z}}\) is the quantized version of the text-aligned speech latent \(\mat{Z}\) from the \(\mathrm{Aggregator}\). 
The quantized indices (tokens) \(\vct{q}\) can be easily obtained with the \(\mathrm{VQ}\) module.

\subsection{TASTE-\textit{S} Decoder}
\label{subsec:taste_s_decoder}
Given the aligned speech embedding \(\mat{\hat{Z}}\) and the text tokens \(\vct{\hat{s}}\) extracted from the TASTE-\textit{S} Encoder, we use the TASTE-\textit{S} Decoder to decode the text-speech aligned tokens back to the waveform. 
The Decoder is consisted by the two components: the auto-regressive (AR) \(\mathrm{Unit\_decoder}\) and the flow-matching~\cite{flow4gen, matcha} \(\mathrm{Vocoder}\). 
Their inputs/outputs are:
\begin{align}
    \vct{\hat{y}} = \mathrm{Unit\_decoder}(\vct{\hat{s}}, \mat{\hat{Z}}), \quad X' = \mathrm{Vocoder}(\vct{\hat{y}}), 
\end{align}
where \(\vct{\hat{y}}\) are the predicted target units introduced as an intermediate for fine-grained speech reconstruction, following TASTE. 
We make several modifications to allow the modules in our Decoder streamable, which are primary based on CosyVoice 2~\cite{cosyvoice2}. 
For the \(\mathrm{Vocoder}\), we simply replace the non-causal vocoder with the causal version. 
As for the \(\mathrm{Unit\_decoder}\), we adopt a streamable pattern during training by interleaving the text-algined tokens with the target units by a predefined \(N:M\) ratio. Figure~\ref{fig:taste_s_method} illustrates the interleaving pattern when the ratio \(N:M\) is set to \(1:3\). 
These modifications allow us to generate the speech chunk-by-chunk, which significantly reduces the latency and enables streaming decoding.

\subsection{Objectives and Training Procedure}
\label{subsec:obj_and_training_proc}
We optimize TASTE-\textit{S} with two objectives: \textbf{(1)} training the built-in ASR module and \textbf{(2)} learning reconstruction conditioned on text and text-aligned tokens.
Given an oracle transcription $\vct{s}$, we train the $\mathrm{CTC\_decoder}$ with the CTC loss
\begin{align}
    \mathcal{L}_{\text{CTC}} &= - \log p_{\text{CTC}}(\vct{s} \mid \mat{O}),
\end{align}
where $p_{\text{CTC}}(\vct{s} \mid \mat{O})$ marginalizes over all valid alignments that correspond to $\vct{s}$.
For reconstruction, given the target unit sequence $\vct{y}$, we train the auto-regressive $\mathrm{Unit\_decoder}$ using the cross-entropy loss
\begin{align}
    \mathcal{L}_{\text{CE}} &= - \sum^{T}_{t=1} \log p(\vct{y}_{t} \mid \vct{s}, \mat{Z}, \vct{y}_{<t}),
\end{align}
where $p(\cdot)$ is parameterized by the $\mathrm{Unit\_decoder}$.

Since the ASR predictions are unreliable at the beginning of training, we adopt a two-stage training procedure.
In Stage~I, we train the $\mathrm{CTC\_decoder}$ independently (without using its output to the other modules).
Meanwhile, the $\mathrm{Aggregator}$ and $\mathrm{Unit\_decoder}$ are trained using the oracle transcription $\vct{s}$, and we bypass the $\mathrm{VQ}$ module.
This stage provides an upper-bound performance reference by removing the effects of ASR errors and quantization bottleneck.
In Stage~II, we enable the full tokenizer and optimize the reconstruction objective
\( \mathcal{L}_{\text{joint}}
    = - \sum^{T}_{t=1} \log p(\vct{y}_{t} \mid \vct{\hat{s}}, \mat{\hat{Z}}, \vct{y}_{<t}),
\)
where $\vct{\hat{s}}$ is the CTC prediction and $\mat{\hat{Z}}$ is the quantized text-aligned speech embedding. 
These design choices are verified in our main results.
 

\section{Experiments}
\label{sec:experiments}

\definecolor{darkred}{RGB}{189,0,0} 
\newcommand{\darkred}[1]{\textcolor{darkred}{#1}}
\begin{table*}[!ht]
    \renewcommand{\arraystretch}{1.0}
    \renewcommand{\cmidrulewidth}{0.1pt}
    \setlength\tabcolsep{4pt}
    \setlength{\aboverulesep}{0.5pt}
    \setlength{\belowrulesep}{0.5pt}
    \caption{
    The main results of our TASTE-\textit{S} Tokenizer. The first section shows the results of the conventional tokenizers; while the second section presents the text-aligned or text-awared methods. We use EXT to denote the reliance on an external ASR; and CTC for using the built-in CTC ASR for TASTE\textit{S}. We use gray row color to indicate the conventional tokenizers with matched bitrates; while the dark-red text indicates that streaming is not feasible. *: Estimated from the throughput of a 13B LLM measured using the vLLM engine~\cite{vllm}.
    }
    \centering
    \begin{adjustbox}{width=2\columnwidth, center}
    \begin{tabular}{l rrr | rccc | cc | c }
    \toprule
    \textbf{Method} & \textbf{Freq.} & \textbf{Bitrate} & \textbf{ASR} & \textbf{WER} & \textbf{UTMOS\(\uparrow\)} & \textbf{Spkr. Sim.\(\uparrow\)}  & \textbf{Drtn. Con.\(\uparrow\)} &  \textbf{Enc. RTF} & \textbf{Dec. RTF} & \textbf{SLM RTF*}  \\
    \toprule
    Original Waveform    & 16k & 256k & N/A & 2.1\% & 4.09 & - & - & - & - & - \\
    \midrule
    \midrule
    \rowcolor{gray!15}
    \multirow{2}{*}{\cellcolor{white}{Encodec~\cite{encodec}}} 
    & 75 & 1500 & N/A & 5.1\% & 1.58 & 0.63 & 0.905 & 0.004 & 0.002 & \darkred{1.524} \\
    & 75 & 3000 & N/A & 2.6\% & 2.35 & 0.78 & 0.928 & 0.004 & 0.002 & \darkred{1.524}\\
    \cmidrule(lr){1-11}
    \rowcolor{gray!15}
    \multirow{2}{*}{\cellcolor{white}{SpeechTokenizer~\cite{speechtokenizer}}} 
    & 50 & 500  & N/A & 5.2\% & 1.27 & 0.35 & 0.916 & 0.004 & 0.003 & \darkred{1.016} \\
    & 50 & 2000 & N/A & 3.0\% & 3.56 & 0.80 & 0.935 & 0.004 & 0.003 & \darkred{1.016} \\
    \cmidrule(lr){1-11}
    \rowcolor{gray!15}
    \multirow{2}{*}{\cellcolor{white}{DM-Codec~\cite{dmcodec}}} 
    & 50 & 1000 & N/A & 10.3\% & 2.42 & 0.49 & 0.896 & 0.008 & 0.006 & \darkred{1.016} \\
    & 50 & 4000 & N/A & 2.5\%  & 3.46 & 0.73 & 0.963 & 0.008 & 0.006 & \darkred{1.016} \\ 
    \cmidrule(lr){1-11}
    \rowcolor{gray!15}
    \cellcolor{white}{BigCodec~\cite{bigcodec}}
    & 80 & 1040 & N/A & 3.0\%  & 4.11 & 0.91 & 0.962 & 0.008 & 0.012 & \darkred{1.626}\\ 
    \cmidrule(lr){1-11}
    \rowcolor{gray!15}
    \cellcolor{white}{WavTokenizer~\cite{wavtokenizer}}
    & 40 & 480 & N/A  & 11.3\%  & 3.57 & 0.71 & 0.925 & 0.003 & 0.001 & 0.813 \\ 
    \cmidrule(lr){1-11}
    \rowcolor{gray!15}
    \cellcolor{white}{Mimi~\cite{moshi}}
    & 12.5 & 1000 & N/A & 3.1\% & 3.60 & 0.83 & 0.928 & 0.002 & 0.001 & 0.254 \\
    \midrule
    \midrule
    TaDiCodec~\cite{tadicodec}
    & 6.25 & 87.5 & EXT & 5.1\% & 3.96 & 0.81 & 0.843 & 0.117 & 0.315 & 0.127 \\
    TaDiCodec (no VQ) 
    & 6.25 & \(>\)1000 & EXT & 4.9\% & 3.99 & 0.82 & 0.869 & 0.117 & 0.315 & 0.127 \\
    \cmidrule(lr){1-11}
    \cmidrule(lr){1-11}
    Text-only
    & \(\sim\)3 & \(\sim\)50 & EXT & 5.7\% & 4.28 & 0.78  & 0.476 & 0.116 & 0.414 & 0.061 \\
    \cmidrule(lr){1-11}
    TASTE~\cite{taste}
    & \(\sim\)3 & \(\sim\)150 & EXT & 4.5\% & 4.24 & 0.80 & 0.844 & 0.117 & 0.414 & 0.061 \\
    \cmidrule(lr){1-11}
    TASTE-\textit{S} (no bi-stage)
    & \(\sim\)3 & \(\sim\)600 & EXT & 10.8\% & 4.01 & 0.72 & 0.798 & 0.117 & 0.076 & 0.061 \\
    \cmidrule(lr){1-11}
    \multirow{2}{*}{TASTE-\textit{S} (no joint)}
    & \(\sim\)3 & \(\sim\)600 & EXT & 4.1\% & 4.11 & 0.88 & 0.900 & 0.117 & 0.076 & 0.061 \\
    & \(\sim\)3 & \(\sim\)600 & CTC & 4.5\% & 4.11 & 0.88 & 0.901 & \textbf{0.002} & 0.076 & 0.061 \\
    \cmidrule(lr){1-11}
    \multirow{2}{*}{TASTE-\textit{S}}
    & \(\sim\)3 & \(\sim\)600 & EXT & 3.9\% & 4.12 & 0.88 & 0.900 & 0.117 & 0.076 & 0.061 \\
    & \(\sim\)3 & \(\sim\)600 & CTC & 4.1\% & 4.11 & 0.88 & 0.901 & \textbf{0.002} & 0.076 & 0.061 \\
    \bottomrule
    \end{tabular}
    \end{adjustbox}
    \label{tab:taste_s_main}
\end{table*}

\subsection{Setup}
\label{subsec:setup}

\noindent\textbf{Dataset} \quad 
We use a random English subset of Emilia~\cite{emilia} (\(\sim\)400 hours) and the full training set from LibriTTS~\cite{libritts} (\(\sim\)600 hours) as our training data. 
Following TASTE, the \textit{test-clean} split from LibriSpeech~\cite{librispeech} is preserved for evaluation. 

\noindent\textbf{Models and Training Configurations} \quad The \(\mathrm{ASR\_encoder}\), \(\mathrm{CTC\_decoder}\), and \(\mathrm{Aggregator}\) are initialized from Whisper~\cite{whisper}.
The \(\mathrm{Unit\_decoder}\) and the \(\mathrm{Vocoder}\) are initialized from CosyVoice 2~\cite{cosyvoice2}. 
During training, the \(\mathrm{ASR\_encoder}\) and the \(\mathrm{Vocoder}\) are frozen. 
In each training stage, the learning rate is set to \texttt{1e-3} for \(\mathrm{CTC\_decoder}\) and \texttt{2e-4} for the other submodules. 
We use \(\mathrm{AdamW}\)~\cite{adamw} optimizer, and a cosine-decay scheduler is used, with linear warmup steps set to \(2000\). 
We use batch size \(128\) and train each stage over \(5\) epochs. 
We set the interleaving ratio \(N:M=2:5\) for the streaming decoding. 
The whole bi-stage training takes around one day to finish on 2 NVIDIA H100 GPUs.

\subsubsection{Metrics}
\label{subsubsec:metrics}
We evaluate our tokenizer from two perspectives: \textbf{(1) reconstruction} and \textbf{(2) streaming efficiency}.

\noindent\textbf{Reconstruction.} \quad
We measure \textit{quality} with \textbf{WER} and \textbf{UTMOS}.
WER is computed between the original and reconstructed speech using transcriptions from a HuBERT-based ASR system.\footnote{https://huggingface.co/facebook/hubert-large-ls960-ft}
UTMOS~\cite{utmos} is a non-intrusive objective metric that predicts MOS from waveform, reported as the average over all reconstructions.
We measure \textit{similarity} with \textbf{Speaker Similarity} and \textbf{Duration Consistency}.
Speaker Similarity is the cosine similarity between speaker embeddings (TDNN~\cite{ecapa-tdnn} based~\cite{cam++}) extracted from original vs.\ reconstructed speech.
Duration Consistency is computed from word-level alignments obtained by Montreal Forced Aligner (MFA), and we count a word as matched if its duration differs by at most \(40\)ms.

\noindent\textbf{Streaming.} \quad
We report \textbf{Real-Time Factor (RTF)} and \textbf{First Chunk Latency (FCL)}.
RTF is defined as $\mathrm{RTF}=\frac{t_{\text{proc}}}{t_{\text{audio}}}$, where $\mathrm{RTF}>1$ indicates slower-than-real-time processing under the measured setup.
FCL is the wall-clock time from when the first audio chunk is available to when the first output chunk (reconstructed speech or text-aligned tokens, depending on the interface) is produced.
All streaming-related metrics are measured on a single NVIDIA A100 GPU.

\subsection{Main Results}
\label{subsec:main_results}
We present the main reconstruction results on LibriSpeech \textit{test-clean} in Table~\ref{tab:taste_s_main}. 
The table is divided into two parts. 
The upper part reports recent conventional speech tokenizers. 
Since these tokenizers take only speech as input, the ASR column is not applicable. 
\textbf{Under matched bitrates (highlighted in gray), TASTE-\textit{S} achieves reconstruction performance comparable to conventional tokenizers.} 
BigCodec is the only conventional tokenizer that consistently attains better reconstruction quality across most metrics. 
However, its tokens are emitted at a very high frequency. 
Modeling it may lead to high latency, making it unsuitable for streaming use cases.
Many conventional tokenizers also share this high-frequency bottleneck.

Next, we turn to the lower part of Table~\ref{tab:taste_s_main}, which reports text-aligned (or text-aware) methods. 
A key practical difference is that all baselines except TASTE-\textit{S} require an external ASR system to obtain text tokens. 
Despite relying only on its built-in ASR, TASTE-\textit{S} achieves competitive or stronger reconstruction quality overall. 
Notably, TASTE attains the highest UTMOS; however, as discussed in the original paper, this result is largely driven by background denoising, which can make the output perceptually cleaner than the source rather than more faithful. 
In TASTE-\textit{S}, reconstruction could better preserve the original perceptual characteristics, yielding a MOS close to that of the input speech. 
In general, \textbf{TASTE-\textit{S} substantially reduces both encoding and decoding RTF, enabling low-latency, streamable tokenization and reconstruction.}

Lastly, we examine the effect of using different ASR systems. 
We observe that TASTE-\textit{S} still delivers strong reconstruction performance when paired with an external ASR system. 
We attribute this stability to the two-stage training in Section~\ref{subsec:obj_and_training_proc}: in the first stage, the decoder is trained with oracle or high-quality pseudo transcriptions, which provides a reliable supervision signal for reconstruction. 
\textbf{Moreover, joint training brings clear benefits.}
In particular, it reduces WER, \textbf{indicating that adapting the decoder to the CTC outputs is important.}
Interestingly, the gain also carries over when using an external ASR system, suggesting improved robustness beyond the specific ASR outputs seen during training.

\subsection{Additional Results and Ablations}

\subsubsection{Longform Speech Reconstruction}
To evaluate the effectiveness of our streamable text-aligned tokenizer, we conduct experiments focusing on longform on-the-fly tokenization and reconstruction. 
The longform speech data is synthesized by concatenating segments from the same speaker in the LibriSpeech \textit{test-clean} set, resulting in 87 audio samples with an average duration of 223.6 seconds. 
We also introduced an additional metric, \textbf{$\Delta$Len}, which measures the average absolute length difference (in \%) between the original and reconstructed audio. 
This metric quantifies the reconstruction accuracy when sequence compression is dynamic. 

The results, presented in Table~\ref{tab:longform}, demonstrate that the WER remains within a reasonable range. 
While TASTE-\textit{S}'s WER is slightly higher than that of the original TASTE, it outperforms the latter in all other metrics, particularly in streaming performance. 
Notably, since the original TASTE decoder is non-causal, its high first-chunk latency makes it unsuitable for real-time applications. 
In contrast, TASTE-\textit{S} employs a causal and streamable decoder, making it highly compatible with streaming scenarios.Interestingly, using the built-in ASR yields the most temporally aligned reconstruction in terms of length difference. 
While employing an external ASR further reduces the WER, it leads to a slight increase in the length difference.

\label{subsec:ablations_and_analysis}
\begin{table}[!t]
    \centering
    \caption{
    The longform speech reconstruction results. W.S indicates the window size of each on-the-fly Encoding/Decoding process. FCL is first chunk latency, as introduced in Section~\ref{subsubsec:metrics}. \(\Delta\)Len is the absolute length difference (in \%). 
    }
    \setlength\tabcolsep{3pt}
    \setlength{\aboverulesep}{0.5pt}
    \setlength{\belowrulesep}{0.5pt}
    \begin{adjustbox}{width=\columnwidth, center}
    \begin{tabular}{l c | ccc | cc}
    \toprule
    \textbf{Method} & \textbf{ASR (W.S.)} & \textbf{WER} & \textbf{\(\Delta\)Len (\%)} & \textbf{Spkr. Sim.\(\uparrow\)} & \textbf{RTF} & \textbf{FCL} \\
    \toprule
    TASTE   & EXT (30s) & 4.2\% &  1.2 \(\pm\) 1.4 & 0.78 & 0.439 & \darkred{12.29} \\
    \midrule
    \midrule
    TASTE-\textit{S} & CTC (10s) & 5.1\% & 1.0 \(\pm\) 0.8 & 0.85 & 0.103 & 0.312 \\
    TASTE-\textit{S} & CTC (15s) & 4.6\% & 0.9 \(\pm\) 0.7 & 0.86 & 0.088 & 0.312 \\   
    TASTE-\textit{S} & CTC (20s) & 4.7\% & 0.9 \(\pm\) 0.9 & 0.87 & 0.081 & 0.311 \\
    TASTE-\textit{S} & CTC (30s) & 4.5\% & \textbf{0.7 \(\pm\) 0.6} & \textbf{0.87} & \textbf{0.081} & \textbf{0.311} \\
    \cmidrule{1-7}
    TASTE-\textit{S} & EXT (30s) & \textbf{3.7\%} & 1.1 \(\pm\) 1.1 & 0.86 & 0.199 & 0.428 \\
    \bottomrule
    \end{tabular}
    \end{adjustbox}
\label{tab:longform}
\end{table}

\begin{table}[!b]
    \centering
    \caption{
    Comparison of TASTE and TASTE-\(S\) across various bitrates and embedding dimensions. 
    }
    \setlength\tabcolsep{3pt}
    \setlength{\aboverulesep}{0.5pt}
    \setlength{\belowrulesep}{0.5pt}
    \begin{adjustbox}{width=\columnwidth, center}
    \begin{tabular}{l rr | cc  cc}
    \toprule
    \textbf{Method} & \textbf{bitrate} & \textbf{emb.} & \textbf{WER\(\downarrow\)} & \textbf{UTMOS} & \textbf{Spkr. Sim.} & \textbf{Drtn. Con.} \\
    \toprule
    TASTE   & \(\sim\)150 & 256 & 4.5\% & 4.24 & 0.80 & 0.844 \\
    \midrule
    \midrule
    TASTE-\textit{S} & \(\sim\)150 & 32 & 4.2\% & 4.13 & 0.86 & 0.857 \\
    TASTE-\textit{S} & \(\sim\)300 & 64 & 4.1\% & 4.12 & 0.88 & 0.887 \\   
    TASTE-\textit{S} & \(\sim\)600 & 128 & 3.9\% & 4.12 & 0.88 & 0.900 \\
    \bottomrule
    \end{tabular}
    \end{adjustbox}
\label{tab:bitrate}
\end{table}

\subsubsection{Ablation Study}

Ablation Study on Bitrate and Quantization. To further investigate the efficiency of TASTE-\textit{S}, we conducted an ablation study across different bitrates, as shown in Table~\ref{tab:bitrate}. It is worth noting that our objective in increasing the bitrate is not merely to boost raw performance. Even when configured with a similar bitrate (\(\sim\)150 bps) as the original TASTE, TASTE-\textit{S} achieves superior WER (\(4.2\%\) vs. \(4.5\%\)) and speaker similarity (\(0.86\) vs. \(0.80\)).

Furthermore, while the bitrate scales up, the embedding dimension in TASTE-\textit{S} remains much lower than that of TASTE (e.g., 32 or 64 vs. 256). This efficiency stems from our adoption of Finite Scalar Quantization (FSQ~\cite{fsq}) as the quantization mechanism. 
Unlike traditional Vector Quantization (VQ) which suffers from codebook utilization issues, FSQ enables high utilization without the need for an explicit codebook, allowing TASTE-\(S\) to maintain high reconstruction fidelity with more compact latent.

\begin{figure}[!t]
    \centering
    \includegraphics[width=\linewidth]{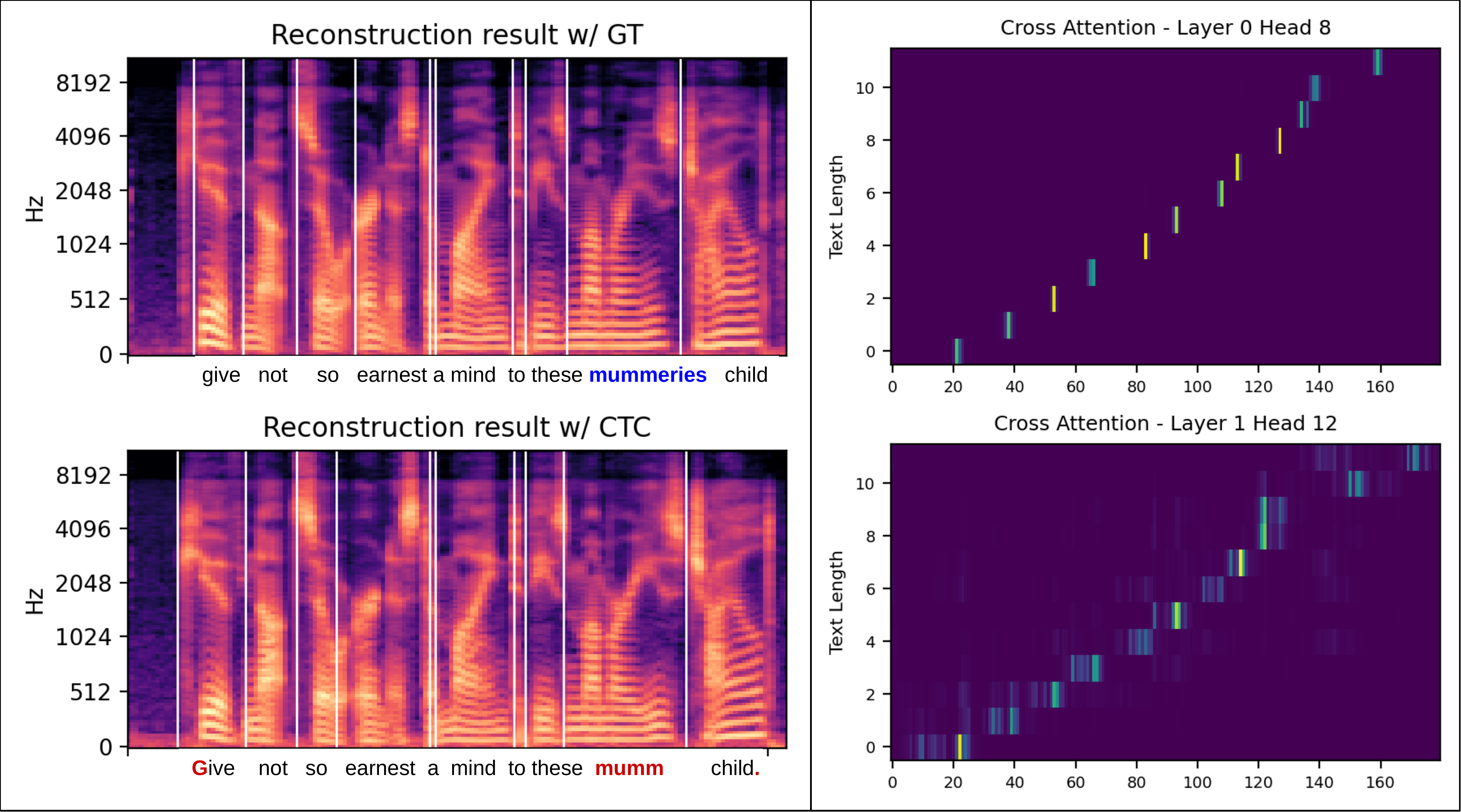}
    \caption{
    Visualization results. Left: reconstruction remains nearly unchanged when using the ground-truth transcript versus the CTC transcript (blue/red mark their mismatched parts). Right: the aggregator shows clear text–speech aligned cross-attention, consistent with TASTE.  
    }
    \label{fig:visualization}
\end{figure}

\subsubsection{Visualization}
Figure~\ref{fig:visualization} provides qualitative evidence for two key behaviors of TASTE-\textit{S}.
On the left, we compare reconstructions produced by conditioning on the ground-truth transcript and on a CTC transcript.
The blue and red tokens highlight where the two transcripts disagree.
Despite these mismatches, the reconstructed spectrograms are almost identical, indicating that the tokenization and reconstruction are robust to imperfect or different transcripts.
On the right, we visualize the cross-attention maps inside the aggregator.
The attention concentrates along a clear diagonal pattern, showing a consistent alignment between text positions and speech frames.
This text--speech aligned behavior matches what is observe in the original TASTE, suggesting that our streamable design preserves the same alignment property.

\section{Conclusion}
We presented TASTE-\textit{S}, a streamable extension of the TASTE framework for text-aligned speech tokenization and embedding in joint spoken language modeling.
By integrating a lightweight CTC-based ASR module into the encoder and redesigning the decoder to be fully causal with chunk-wise generation, TASTE-\textit{S} removes the reliance on external offline ASR and enables low-latency, on-the-fly reconstruction.
Experiments show that TASTE-\textit{S} matches or improves upon TASTE in reconstruction fidelity while substantially reducing both encoding and decoding RTFs, making streaming deployment practical.
Further analyses demonstrate effective long-form reconstruction with accurate temporal consistency.
The visualizations further confirm robust reconstruction under imperfect transcripts and clear text--speech aligned cross-attention in the aggregator.
Overall, TASTE-\textit{S} enables low-latency, text-aligned speech tokenization and reconstruction, making it a practical component for building responsive and semantically grounded joint SLMs.

\section{Generative AI Use Disclosure}
We use generative AI tools to assist with language editing (e.g., polishing phrasing, improving clarity, and correcting grammar) during manuscript preparation. 
All technical content, experimental design, results, and conclusions are produced by the authors, who take full responsibility for the paper. 
No generative AI tool is listed as an author, and the tool is not used to generate a significant portion of the manuscript.

\bibliographystyle{IEEEtran}
\bibliography{mybib}

\end{document}